%% file: main.tex
 \documentclass[journal]{IEEEtran}
\usepackage[utf8]{inputenc}
\usepackage{siunitx}
\usepackage{comment}
\usepackage{amssymb}
\usepackage{amsmath,pict2e}
\usepackage{gensymb}
\usepackage{graphicx}
\usepackage{graphics}
\usepackage{times}
\usepackage{xcolor}
\usepackage{xspace}
\usepackage{textcomp}
\usepackage{bm}
\usepackage{diagbox}
\usepackage[mathscr]{euscript} 
\usepackage{multirow}
\usepackage{epstopdf}
\usepackage{multicol}
\usepackage{hyperref}
\usepackage{mathtools}

\newcommand{\edd}[1]{\textcolor{black}{#1}}

\input{macro} 
\newcommand{\myurl}{http://web.eecs.umich.edu/~fessler/irt/irt}

\begin{document}
%
\title{Sparse-view Cone Beam CT Reconstruction using Data-consistent Supervised and Adversarial Learning from Scarce Training Data}


\author{\IEEEauthorblockN{%
Anish Lahiri\IEEEauthorrefmark{1},
Marc L. Klasky\IEEEauthorrefmark{2},
Jeffrey A. Fessler\IEEEauthorrefmark{1} \IEEEmembership{Fellow,~IEEE},
and
Saiprasad Ravishankar\IEEEauthorrefmark{3} \textit{Senior Member, IEEE}\\}
\thanks{\IEEEauthorrefmark{1}%
A. Lahiri and J. A. Fessler are with the Department of Electrical and Computer Engineering, University of Michigan, Ann Arbor, MI 48014 USA.
Emails: anishl@umich.edu, fessler@umich.edu. \\
\IEEEauthorrefmark{3}%
S. Ravishankar is with the Department of Computational Mathematics, Science and Engineering, and the Department of Biomedical Engineering, Michigan State University, East Lansing, MI 48824 USA. Email: ravisha3@msu.edu.\\
\IEEEauthorrefmark{2}%
M. L. Klaksy is with the Theoretical Division, Los Alamos National Laboratory, Los Alamos, NM 87545 USA. Email: mklasky@lanl.gov.
}}

%

\IEEEtitleabstractindextext{%
\begin{abstract}
\input{abs}
\end{abstract}

}


\maketitle

\IEEEdisplaynontitleabstractindextext

%
\IEEEpeerreviewmaketitle

\section{Introduction}
\label{intro}
\input{s,intro}

\section{Algorithm \& Problem Setup}
\label{algo}
\input{s,algo}

\section{Methods}
\label{methods}
\input{s,methods}

\section{Results}
\label{results}
\input{s,results}

\section{Discussion}
\label{disc}
\input{s,disc}

\section{Conclusions and Future Work}
\label{conc}
\input{s,conc}

\bibliographystyle{IEEEtran}
\input{main.bbl}



\end{document}

%% file: macro.tex
\newcommand{\fref}[1] {Fig.~\ref{#1}\xspace}

\newcommand{\tref}[1] {Table~\ref{#1}\xspace}

\newcommand{\argmin}[1] {\underset{#1}{\textnormal{arg min}}}

\newcommand{\xmath}[1] {\ensuremath{#1}\xspace}
\newcommand{\blmath}[1] {\xmath{\bm{#1}}}

\newcommand{\x}{\xmath{\bm{x}}}

\newcommand{\y}{\xmath{\bm{y}}}

\newcommand{\A}{\blmath{A}}

\newcommand{\respp}[1]{\marginpar{\textcolor{blue}{}}}
\newcommand{\resp}[1]{\marginpar{\textcolor{blue}{}}}

\makeatletter
\newcommand{\bigcomp}{%
  \DOTSB
  \mathop{\vphantom{\sum}\mathpalette\bigcomp@\relax}%
  \slimits@
}
\newcommand{\bigcomp@}[2]{%
  \begingroup\m@th
  \sbox\z@{$#1\sum$}%
  \setlength{\unitlength}{0.9\dimexpr\ht\z@+\dp\z@}%
  \vcenter{\hbox{%
    \begin{picture}(1,1)
    \bigcomp@linethickness{#1}
    \put(0.5,0.5){\circle{1}}
    \end{picture}%
  }}%
  \endgroup
}
\newcommand{\bigcomp@linethickness}[1]{%
  \linethickness{%
      \ifx#1\displaystyle 2\fontdimen8\textfont\else
      \ifx#1\textstyle 1.65\fontdimen8\textfont\else
      \ifx#1\scriptstyle 1.65\fontdimen8\scriptfont\else
      1.65\fontdimen8\scriptscriptfont\fi\fi\fi 3
  }%
}
\makeatother

%% file: abs.tex
Reconstruction of CT images from a limited set of projections through an object is important in 
several applications ranging from medical imaging to industrial settings.
As the number of available projections decreases,
traditional reconstruction techniques
such as the FDK algorithm
and model-based iterative reconstruction methods perform poorly.
Recently, data-driven methods 
such as deep learning-based reconstruction
have garnered a lot of attention in 
applications
because they yield better performance when enough training data is available.
However, even these methods have their limitations
when there is a scarcity of available training data.
This work focuses on image reconstruction in such settings,
i.e., when both the number of available CT projections 
and the training data is extremely limited.
We adopt a 
sequential
reconstruction approach
over several stages
using an adversarially trained shallow network for `destreaking'
followed by a data-consistency update in each stage.
To deal with the challenge of limited data,
we use image subvolumes to train our method,
and patch aggregation during testing.
To deal with the computational challenge of learning on 3D datasets for 3D reconstruction,
we use a hybrid 3D-to-2D mapping network for the `destreaking' part.
Comparisons to other methods over several test examples
indicate that the proposed method
has much potential, 
when both the number of projections 
and available training data are highly limited.

%% file: s,intro.tex

Computed Tomography (CT) 
is an important
imaging modality across 
applications in medicine, industry, science and security.
In this work, we develop an iterative machine learning-based approach
for 3D cone beam CT reconstruction from very limited measurements or projections,
and using limited training data.
In the following, we first review some background in limited-view CT reconstruction 
before highlighting the contributions of this work.

\subsection{Background}

Cone Beam CT (CBCT) is a CT-based technique that allows for 
three-dimensional imaging of an object
using X-rays diverging from a source.
In CBCT, an entire 3D image volume is reconstructed from a set of 2D projections 
through the corresponding object.
These projections/measurements are obtained at 
different angles or `views' around the object,
and are collectively dubbed a \textit{sinogram}.
There are several approaches
for the inverse problem
of obtaining an image from these measurements.
A classical method for this task is the analytical Feldkamp-Davis-Kress (FDK) algorithm~\cite{feldkamp:84:pcb}.
More sophisticated methods for 2D or 3D reconstruction involve model-based reconstruction
using iterative algorithms
\cite{thibault:07:atd,xu:12:ias,cho:15:rdf},
and data-driven algorithms
\cite{wang:16:apo,chun:17:svx}.

Model-based image reconstruction (MBIR) or statistical image reconstruction (SIR) methods
exploit sophisticated models for the physics of the imaging system and models
for sensor and noise statistics as well as for the underlying object.
These methods iteratively optimize for the underlying image
based on the system forward model, measurement statistical model,
and assumed prior for the underlying object%
~\cite{elbakri:02:sir,sauer:93:alu,fessler:00:sir,thibault:06:arf}.
In particular, penalized weighted least squares (PWLS) approaches
have been popular for CT image reconstruction
that optimize a combination of a statistically weighted quadratic data-fidelity term
(capturing the forward and noise model)
and a regularizer penalty that captures prior information of the object~\cite{depierro:93:otr}.
MBIR methods have often used simple regularizers~\cite{ravishankar:20:irf}
such as edge-preserving regularization
involving nonquadratic functions of differences between neighboring pixels~\cite{ahn:15:qco}
(implying image gradients may be sparse) or other improved
regularizers~\cite{YU2017808,Xu:20,zhang2020,kim2019}.

Within the class of data-driven approaches,
dictionary learning~\cite{ravishankar:20:irf} and
deep learning based methods for reconstruction have gained popularity in recent years
due to their demonstrated effectiveness
in removing artifacts from images in a variety of modalities,
great flexibility and the availability of curated datasets for training%
~\cite{anirudh:17:ltv,han:18:dlr,kim:19:efv}.

However, in several applications,
acquiring many projections or `views' through the object
may be undesirable or impossible.
This constraint may be to reduce exposure to radiation in medical imaging applications,
or due to only pre-set limited or sparse views being possible
in industrial or security applications.
Moreover, in dynamic imaging applications, where the object is changing while being imaged,
we would also be limited to fewer views per temporal state, to prevent blurring.  
While total variation (TV)-based MBIR methods have been extensively applied
to such sparse-view and sparse angle reconstruction problems%
~\cite{sidky:06:air,yu:09:csb,bian:10:eos,ramani:12:asb,herman:08:irf,chen:08:pic},
other model-based CT reconstruction algorithms
rely upon learned prior-based regularizers%
~\cite{pfister:14:mbi,zhang:16:ldc,zheng:18:pua}. 
These methods often require many iterations to converge,
leading to large runtimes,
and also require careful selection of regularization parameters 
to obtain reasonable image quality trade-offs.  

Deep learning-based algorithms have also found considerable use in many problems
ranging from artifact correction
 to combination with model-based reconstruction%
~\cite{chen:18:lle,wu:17:ild,chun:18:fac,chun:19:mnf-arxiv}.
Deep learning approaches could be supervised or unsupervised or mixed~\cite{superct21,lahiri:21:bps}
and include
image-domain (denoising) methods, sensor-domain methods, AUTOMAP,
as well as hybrid-domain methods
(cf. reviews in~\cite{ravishankar:20:irf,ongie_deep_2020}).
Hybrid-domain methods are gaining increasing interest
and enforce data consistency
(i.e., the reconstruction should be consistent with the measurement model)
during training and reconstruction to improve stability and performance.
Deep learning methods often require large training data sets and long training times to work well.
They may also struggle to generalize to data with novel features
or that are obtained with different experimental settings.

When reconstructing 3D objects from extremely limited tomographic views or projections,
many of the aforementioned approaches fail.
The FDK algorithm yields reconstructions that are severely ridden with streak artifacts.
While conventional iterative methods perform better,
the quality of reconstructions leave a lot to be desired,
and there is often poor bias-variance trade-offs. 
Deep learning-based approaches have the potential to perform better in this scenario,
but still perform poorly when there is a scarcity of available data for training,
such as in national security applications
where experimental data is limited
and accurate simulations are expensive
\cite{hossain:21:hpi}.
While there are approaches that reconstruct from very limited projections,
they either 
do not target 3D CBCT imaging~\cite{han:18:dlr,herman:08:irf,quinto:98:eal,guan:20:lva},
or rely upon many paired training image volumes \cite{xiang:20:dlb,anirudh:17:ltv,kim:19:efv}.

\subsection{Contributions}

This paper focuses on developing a method
that can provide quality reconstructions
from extremely limited projections
even with very extremely limited training data.
The proposed reconstruction approach works across multiple stages,
similar to an unrolled-loop algorithm~\cite{ravishankar:20:irf},
where each stage consists of a shallow CNN block trained
using a combined supervised and adversarial loss,
followed by a data-consistency block.
The adversarial component of our loss
yields destreaked images that have more realistic texture. 
To mitigate the challenge of reduced training data,
we reduce the scope of our learning
to patches or image sub-volumes.
This approach allows us to provide several training examples
from even a single training image.
Destreaked patches are aggregated before data-consistency is applied to the whole volume.
Furthermore,
we prime our method using an edge-preserving regularized reconstruction as input. 

We compare our methods to a variety of techniques
including the FDK algorithm,
edge-preserving regularized reconstruction,
and deep CNN-based reconstruction without data-consistency.
Simulation results suggest that the proposed method
provides much better image quality
than previous techniques with extremely limited (four or eight) views of 3D objects.

\subsection{Organization}

The rest of this paper is organized as follows.
Section~\ref{algo} describes the proposed approach in detail.
Section~\ref{methods} explains our choices for various algorithm parameters
as well as our experimental setup.
Section~\ref{results} presents the results of our comparisons to other algorithms
as well as other experiments that offer insights into
the process of our reconstruction.
Section~\ref{disc} elaborates upon these observations.
Finally, Section~\ref{conc} states our conclusions and offers some avenues for future research.

%% file: s,algo.tex
\begin{figure*}[h!]
    \centering
    \includegraphics[width=0.95\textwidth]{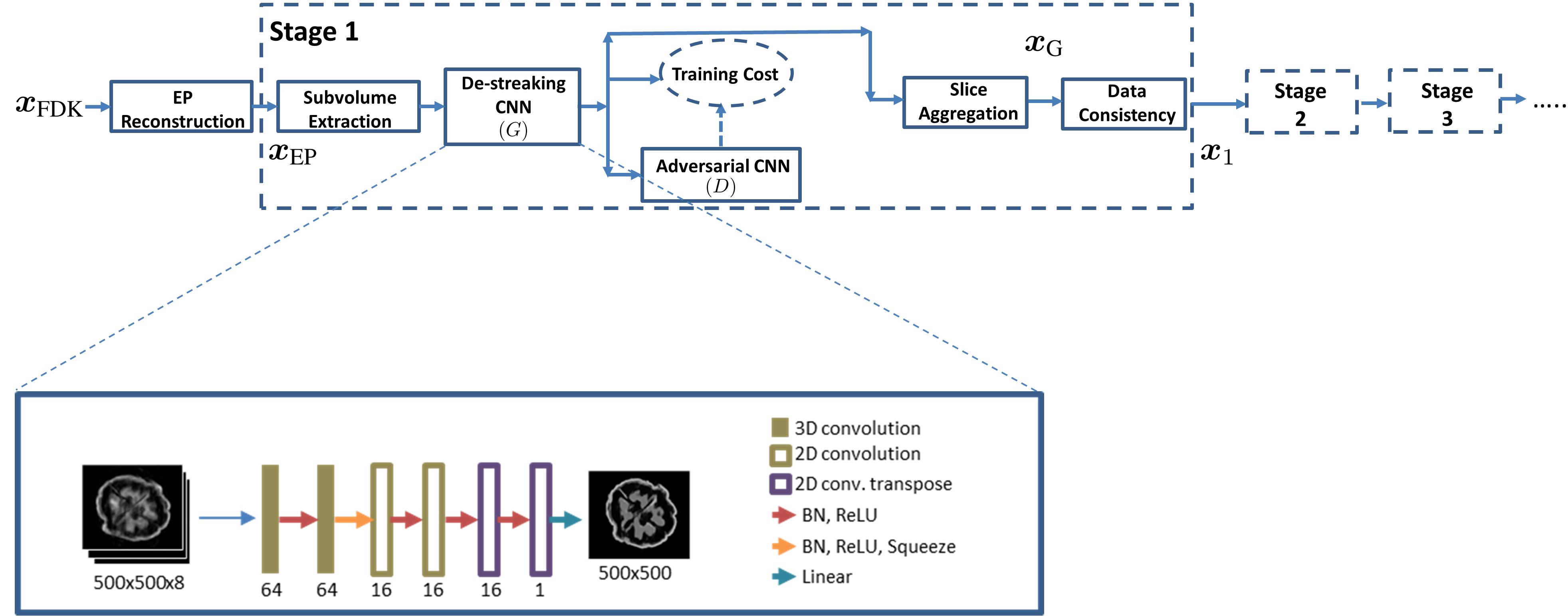}
    \caption{Flow diagram depicting the overall pipeline of our algorithm, where $\bm{x}_\text{FDK}$ is the FDK reconstruction, $\bm{x}_\text{EP}$ is an edge-preserving regularized reconstruction, $\bm{x}_\text{G}$ is the generator's output after slice aggregation, and $\bm{x}_1$ is the output of the first stage.}
    \label{fig:pipeline}
\end{figure*}

Our proposed method for CBCT reconstruction
focuses on addressing two primary challenges:
a very limited number of available views,
and limited number of available training objects.
We address the former through a combination of three aspects:
(1) using an edge-preserving regularized reconstruction~\cite{cho:15:rdf}
to initialize our iterative-type algorithm;
(2) including an adversarial component to the training loss function
for our learned destreaking networks
(similar to generative adversarial networks or GANs); and
(3) including data-consistency blocks
that reinforce acquired measurements in the destreaked 3D reconstruction. 

The problem of scarce training data is addressed primarily
by two approaches.
First, we split an entire image volume into patches in the form of overlapping subvolumes.
Essentially, this step localizes the scope of CNN-based destreaking
to a comparatively smaller neighborhood,
while allowing us to generate many training examples from a single image volume. 
Second, we use a shallow destreaking CNN
to avoid overfitting to the training data. 
To reduce computation time associated with multiple 3D convolutions
and subsequent patch aggregation,
the CNN is designed to map 3D subvolumes to 2D slices
\cite{ziabari:18:2dl}. 
This approach enables using
3D contextual information for the destreaking task,
while removing the need for patch averaging (of overlapping 3D patches)
and associated artifacts during aggregation. 

Many learning-based reconstruction approaches
in applications like MRI
work through end-to-end training,
whereas such approaches
are less practicle in CT
due to the complexity of the system matrix. 
Thus, our developed algorithm operates as a multi-stage greedy approach
similar to works like~\cite{lim:20:ilc,cordadincan:21:met}.
Each stage is composed of a CNN that maps each 3D subvolume with streaking artifacts
to a clean 2D slice corresponding to the slice at the centre of the subvolume.
Because the objects considered here have finite support,
we treat the slices at the edge of the volume
in the direction of aggregation are being empty,
and set them to zero.
One could use other boundary conditions
for long objects
\cite{magnusson:06:hol}.
Once an entire image volume has been aggregated from individual clean slices,
this volume is passed through a data-consistency update
to reinforce acquired measurements
and reduce any `hallucinations' introduced by the network.
This output subvolume is then provided as input to the next stage.
\fref{fig:pipeline} depicts the process
that is akin to algorithm unrolling
\cite{monga:21:aui}.
As mentioned earlier,
to reduce noise and streaks,
the input to the first stage of our method
is an edge-preserving regularized (iteratively obtained) reconstruction,
using the regularizer in \cite{cho:15:rdf}
and the algorithm in \cite{nien:15:rla},
which in turn is initialized with an FDK reconstruction for faster convergence.

We train the CNN parameters separately for each stage.
The training loss for the destreaking CNN in each stage
consists of a weighted combination of a masked mean squared error term
calculated over a region of interest using the ground truth training image slices,
and an adversarial input from another CNN
that acts as a discriminator for the output of the destreaking CNN
(also specific to the stage).
Adversarial training often is posed as a min-max optimization problem
\cite{goodfellow:14:gan},
but the practical implementation
involves alternating between updating
generator (destreaking) network $G$ parameters $\phi^k$
and discriminator network $D$ parameters $\theta^k$,
with more frequent updates of the generator parameters.
Our approach to updating the generator weights
for the $k$th stage
is mathematically expressed as:
\begin{align}
\begin{split}
  \hat{\theta}^k = \arg\min_{\theta} & \ 
  {-\lambda} \, \mathbb{E}\big[D_{\phi^k}(G_{\theta}(\mathcal{P}_3\x_{k-1}))\big]  \\
  &+ \mathbb{E}\big[\|\mathcal{P}_{3}^{2,\text{mid}}\x_{\text{GT}}-G_{\theta}(\mathcal{P}_3\x_{k-1})\|_2^2\big], 
\end{split}
\end{align}
where
$\x_{k-1}$ is the output of the $(k-1)$th stage of our algorithm,
$\x_0$ is set to be $\x_{\text{EP}}$
(the edge-preserving regularized reconstruction),
$\x_{\text{GT}}$ is the ground truth, 
$\lambda$ is a regularization parameter
that varies as the weights of $G^k$ are updated
(see Section \ref{methods:hyp}),
$\mathcal{P}_3$ is a 3D patch or subvolume extraction operator,
and $\mathcal{P}_3^{2,\text{mid}}$ is an operator
that extracts the 2D central slice from an image subvolume,
where the position of the subvolume is determined by $\mathcal{P}_3$.
\edd{We restricted the reconstructions to a region in the 
image volume containing the object of interest.}
The expectation $\mathbb{E}$
is taken over
the set of training examples. 

Our approach to updating the 
the discriminator network parameters
is likewise given as:
\begin{align}
    \begin{split}
        \hat{\phi}^k = \arg\min_{\phi}~
        &\mathbb{E}\big[\big(D_\phi(G_{\theta^k}(\mathcal{P}_3\x_{k-1}))-0\big)^2\big]\\
        &+ \mathbb{E}\big[\big(D_\phi(\mathcal{P}_{3}^{2,\text{mid}}\x_{\text{GT}})-1\big)^2\big].
    \end{split}
\end{align}
The data-consistency update
involves seeking an image that is consistent with the acquired measurements
while still being `close' to the slice-aggregated destreaked image.
The optimization problem for this step is framed as:
\begin{align}
\label{eq:rls}
    \x_{k}
    = \argmin{\x} \|\A\x-\y\|_2^2 + \beta\|\x - \x_{k,\text{G}}\|_2^2,
\end{align}
where
\A is the CBCT system matrix,
implemented with the separable-footprint projector
\cite{long:10:3fa},
\y denotes the projections or acquired measurements,
$\beta>0$ is a regularization parameter,
and
$\x_{k,\text{G}}$
is the output of the generator
after slice aggregation
at the $k$th stage.
We used an ordinary least-squares (LS)
data-fit term rather than a weighted LS (WLS) term
because the focus here is on sparse views
rather than low-dose imaging,
but the method generalizes directly
to the WLS case.
We used 50 conjugate gradient (CG) iterations
to minimize \eqref{eq:rls}.

%% file: s,methods.tex
\subsection{Dataset and Experimental Settings}
To train and test our method,
we used the publicly available 3D walnut 
CT dataset \cite{sarkissian:19:acb}.
To study the ability to learn from very limited data,
we used a single walnut for training our method,
and tested our 
algorithm on 5 different walnuts.
Furthermore, extremely limited data with
8 or 4 views/projections through the walnuts were
used in training and testing the network.
Separate networks were trained for reconstructing image volumes from
4 and 8 views, respectively.
These CBCT
views were generated using the MIRT~\cite{fessler:16:irt} package,
and were equally spaced over 360 degrees.
The distance from source
to detector was set to be 20 cm, and the 
distance from the object to the detector was 4.08 cm,
and each projection view was
$150 \times 150$ pixels of size $\approx 0.4$~mm square. 
Because the CBCT system simulated here
has a small cone angle
(almost parallel beam),
8 views over 360$^\circ$
probably has only a bit more information
than 4 views over $180^\circ$.
The image volume for each 
walnut was $501\times 501 \times 501$.
The dimensions of each voxel were approx. $0.12 \times 0.12 \times 0.12$ mm$^3$.

\subsection{Hyperparameters and Network Architectures}\label{methods:hyp}

The subvolume
size for our scheme was chosen to be $500 \times 500 \times 8$.
The parameter $\beta$ was chosen to be 1, and $\lambda$ was
changed dynamically as the network weights 
were updated according to $10^{\lfloor\log(r)\rfloor}$,
where $r=\mathbb{E}\big[\|\mathcal{P}_{3}^{2,\text{mid}}\x_{\text{GT}}-G_{\theta}(\mathcal{P}_3\x_{k-1})\|_2^2\big]$.
\edd{This was done to maintain balance between the $\ell_2$ and adversarial 
loss components during generator training.}
The number of stages in our method was set to 4, and the networks in each stage were trained
for 40 epochs. The weights for the discriminator were updated once for every
10 times the weights of the destreaking CNN were updated. 

\fref{fig:pipeline} depicts
the generator architecture we used.
The kernel size for 3D convolutions 
was ($3\times 3 \times 3$), and ($3\times 3$) for the 2D convolutions.
\edd{The discriminator was akin to a classifier with two convolutional layers
and with 8 filters in each convolutional layer
with ($3\times 3$) kernel size with stride 1 followed by fully connected layers
with (1152,8,8) 
nodes respectively, with a sigmoid activation at the final output to constrain the output to be between 0 and 1.} The batch size during training the destreaking CNNs was 6.  
Training the destreaking CNN in each stage of our algorithm took approx. 10
hours on 3 NVIDIA Quadro RTX 5000 GPUs, while at test time, each walnut 
volume required 7 minutes to reconstruct with a batch size of 3 on two of the same GPUs. The data 
consistency update required an additional 3 minutes on a workstation with Intel(R) Xeon(R) Silver 4214 CPU @ 2.20GHz with 48 cores.

\subsection{Compared Methods}

\edd{To assess the performance of our method,
we used 4 stages of our  proposed algorithm
to compromise between image quality and runtime.
We compared the output for all 5 test walnuts
to the conventional FDK} reconstruction,
an edge-preserving regularized (MBIR) reconstruction~\cite{cho:15:rdf},
as well as the slice-aggregated output
from a single stage of our destreaking CNN without data consistency.

\subsection{Performance Metrics}

We primarily used the normalized mean absolute error (NMAE)
as a metric for evaluating the performance of various methods. 
The error is evaluated over the voxels within the region-of-interest (ROI)
of a three-dimensional mask
obtained by dilating 
a ground truth segmentation of the walnut being reconstructed.
The masked region includes all voxels 
within the shell of the walnut.
The NMAE normalization used the mean intensity
of the ground truth voxels within this mask.
Essentially,
$E_{\text{NMAE}}(x_{GT},x_O,\mathscr{M})
=
\|\mathscr{M}\odot (x_{GT}-x_O)\|_1/\|\mathscr{M}\odot x_{GT}\|_1$,
where $x_{GT}$ is the ground truth image volume,
$x_O$ is the reconstruction whose quality is being evaluated,
and $\mathscr{M}$ \edd{is a binary mask specific to the test walnut,
which excludes any pixel not within the a dilation of the outer boundary of the walnut. 
These are obtained by a histogram-based thresholding
of the corresponding ground truth volumes for the test walnuts.}

Another metric that is used for comparison in our work
is the normalized high-frequency error norm or NHFEN
\cite{ravishankar:11:mir}.
We computed the HFEN
for every slice of the reconstructed walnut
as the $\ell_2$ norm of the difference of masked edges (obtained through a high-pass filtering)
between the input and reference images. 
The masking is done similarly as described earlier.
A Laplacian of Gaussian (LoG) filter was used as the edge detector. 
The kernel size was set to $15\times 15$, with a standard deviation of 1.5 pixels. 
The normalization was performed over the high frequency components
of the ground truth image over the masked ROI.
Mathematically, this metric is calculated as
$E_{\text{NHFEN}}(x_{GT},x_O,\mathscr{M})
=
\frac{1}{N} \sum_{i}\|\mathscr{H}(\mathscr{M}[:,:,i]\odot x_{GT}[:,:,i])
-
(\mathscr{H}(\mathscr{M}[:,:,i]\odot x_O[:,:,i])\|_2
/
\|\mathscr{H}(\mathscr{M}[:,:,i]\odot x_{GT}[:,:,i])\|_2$,
where $\mathscr{H}$ denotes the LoG filter described earlier, $i$ 
indexes the slices of the image volume in the $z$ direction, where
$N$ is the total number of slices in that direction, and the other
symbols have their usual meaning, as described previously. 
\edd{An advantage of using such normalized metrics is that it 
allows for the evaluation of the reconstruction quality only in 
areas of interest in the volume,
disregarding the effect of empty spaces around it.}

%% file: s,results.tex

\tref{table:comp_mae_hfen}
compares
the reconstruction performance of various methods (including our own)
described in the previous sections.
The proposed approach
substantially improves the NMAE and NHFEN
compared to the reference methods for both 8 and 4 acquired projections.
As expected, the quality of reconstructions using 4 acquired projections
was worse than when 8 projections were acquired for reconstruction.
     

\begin{table*}[h]
\centering
\begin{tabular}{|c||c|c|c|c|c|c|c|c|}
\hline
\multicolumn{9}{|c|}{\textbf{8 views}}\\
\hline
\textbf{Walnut \#} &\multicolumn{2}{c|}{\textbf{FDK recon. }}&\multicolumn{2}{c|}{\textbf{EP recon. }}&\multicolumn{2}{c|}{\textbf{CNN destreaking }}&\multicolumn{2}{c|}{\textbf{Proposed }}\\\cline{2-9}& NMAE & NHFEN & NMAE & NHFEN & NMAE & NHFEN & NMAE & NHFEN\\\hline 
    1 & 0.77 & 0.90 & 0.45 & 0.58 & 0.40 & 0.59 & \textbf{0.26} & \textbf{0.54}\\\hline
     2 & 0.77 & 0.88 & 0.45 & 0.57 & 0.38 & 0.58 & \textbf{0.25} & \textbf{0.53}\\\hline
     3 & 0.79 & 0.90 & 0.49 & 0.61 & 0.42 & 0.62 & \textbf{0.30} & \textbf{0.58}\\\hline
     4 & 0.79 & 0.96 & 0.45 & 0.62 & 0.39 & 0.62 & \textbf{0.27} & \textbf{0.58}\\\hline
     5 & 0.82 & 0.98 & 0.48 & 0.65 & 0.41 & 0.66 & \textbf{0.27} & \textbf{0.61}\\\hline
\multicolumn{9}{|c|}{\textbf{4 views}}\\
\hline
\textbf{Walnut \#} &\multicolumn{2}{c|}{\textbf{FDK recon. }}&\multicolumn{2}{c|}{\textbf{EP recon. }}&\multicolumn{2}{c|}{\textbf{CNN destreaking }}&\multicolumn{2}{c|}{\textbf{Proposed }}\\\cline{2-9}& NMAE & NHFEN & NMAE & NHFEN & NMAE & NHFEN & NMAE & NHFEN\\\hline 
    1 & 1.08 & 1.11 & 0.62 & 0.61 & 0.61 & 0.61 & \textbf{0.44} & \textbf{0.60}\\\hline
     2 & 1.23 & 1.18 & 0.65 & 0.60 & 0.63 & 0.60 & \textbf{0.45} & \textbf{0.60}\\\hline
     3 & 1.22 & 1.15 & 0.67 & 0.64 & 0.66 & 0.64 & \textbf{0.48} & \textbf{0.64}\\\hline
     4 & 1.22 & 1.22 & 0.68 & 0.64 & 0.67 & 0.65 & \textbf{0.46} & \textbf{0.65}\\\hline
     5 & 1.23 & 1.27 & 0.65 & 0.68 & 0.65 & 0.68 & \textbf{0.46} & \textbf{0.68}\\\hline

\end{tabular}
\vspace{0.1in}
\caption{Comparison of the performance of our proposed method against FDK reconstruction, edge-preserving (EP) regularized reconstruction, a single stage of CNN-based destreaking without data-consistency (using the same architecture as in our method) and our proposed multistage reconstruction algorithm for 8 (top) and 4 (bottom) acquired projections. The metrics used for performance are the normalized mean absolute error (NMAE) and the normalized high-frequency error norm (NHFEN).}
\label{table:comp_mae_hfen}
\end{table*}

\fref{fig:recon:wal1}
gives
an example reconstruction for 8 views (walnut 1 in \tref{table:comp_mae_hfen}),
showing central slices through the walnut in all three orientations
(sagittal, coronal and transverse).
The proposed algorithm provides significantly higher quality 
reconstructions than the other methods.
This is particularly evident in the extent to which our 
algorithm is able to restore the finer features of the walnuts, and has fewer artifacts.
\newcommand{\hh}{2.8in} 

\begin{figure*}
\begin{center}
\begin{tabular}{cc}
\textbf{Ground Truth Test Volume} & \textbf{Ground Truth Training Volume}\\ 
\includegraphics[height=\hh]{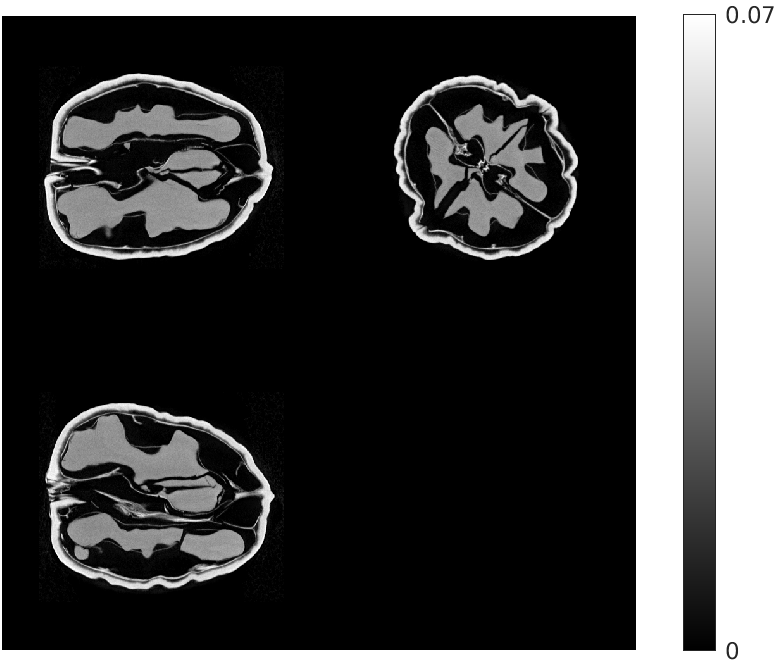} & \includegraphics[height=\hh]{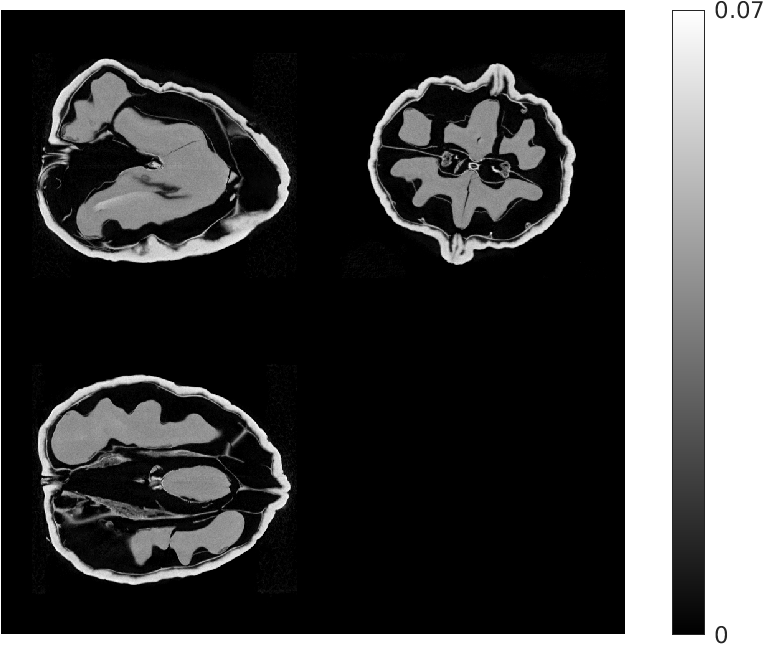} \\
\hspace{0.1in}(a) (NMAE)\hspace{0.1in} & (b) (n/a)\hspace{0.1in}\\
\textbf{FDK} & \textbf{EP Regularized Recon.} \\ 
 \includegraphics[height=\hh]{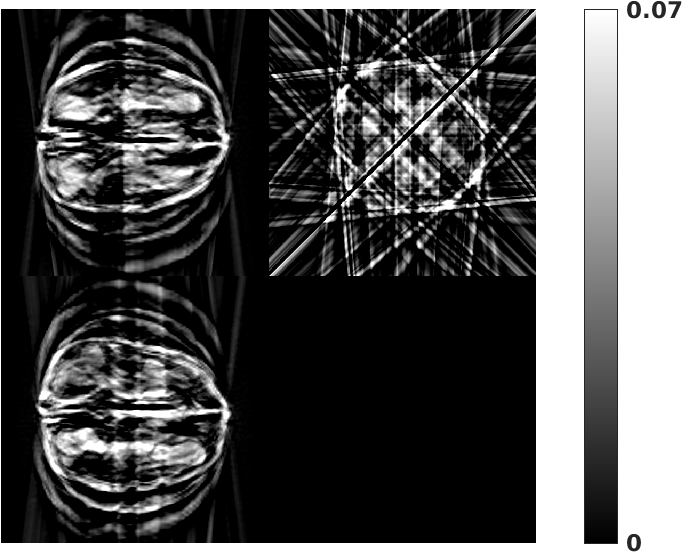}  & \includegraphics[height=\hh]{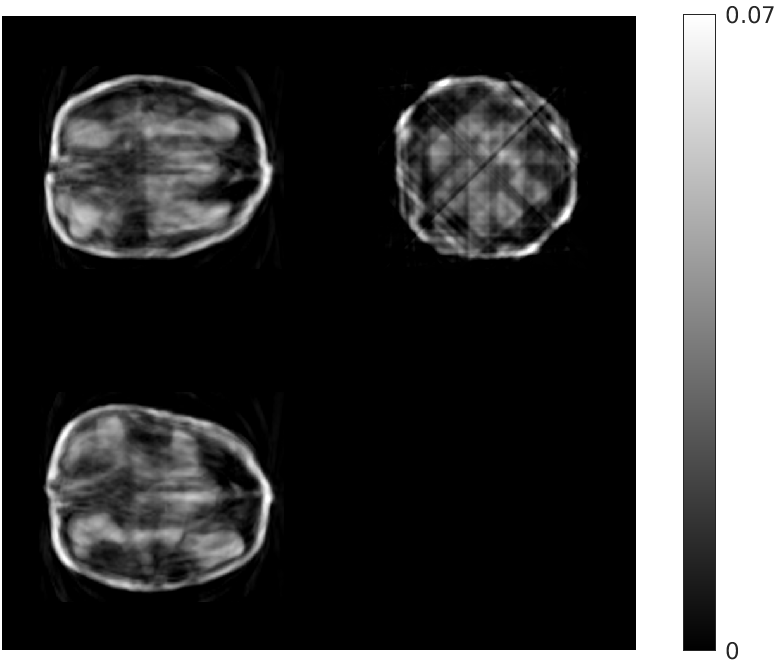}\\
 \hspace{0.1in}(c) (0.77) \hspace{0.1in} & \hspace{0.1in}(d) (0.45)\hspace{0.1in }\\
\textbf{Destreaking CNN} & \textbf{Proposed}\\
 \includegraphics[height=\hh]{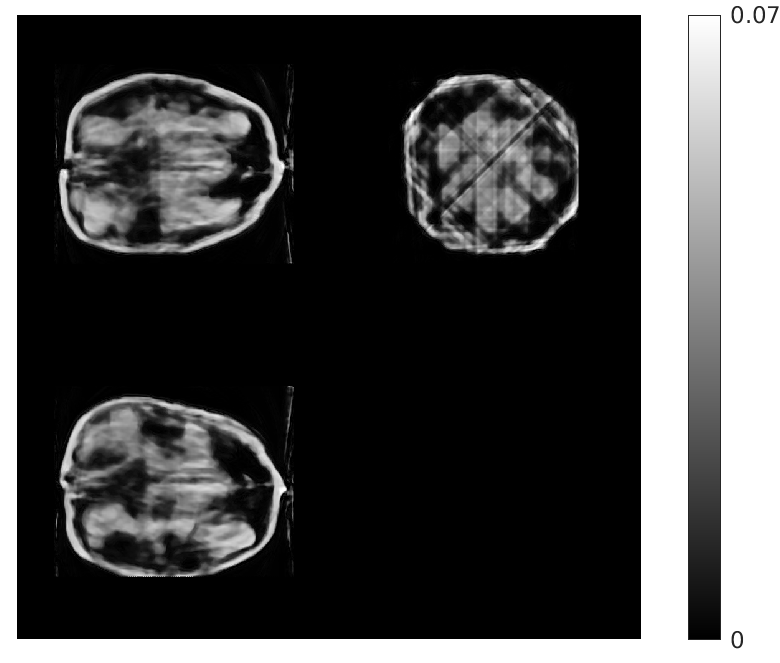}  & \includegraphics[height=\hh]{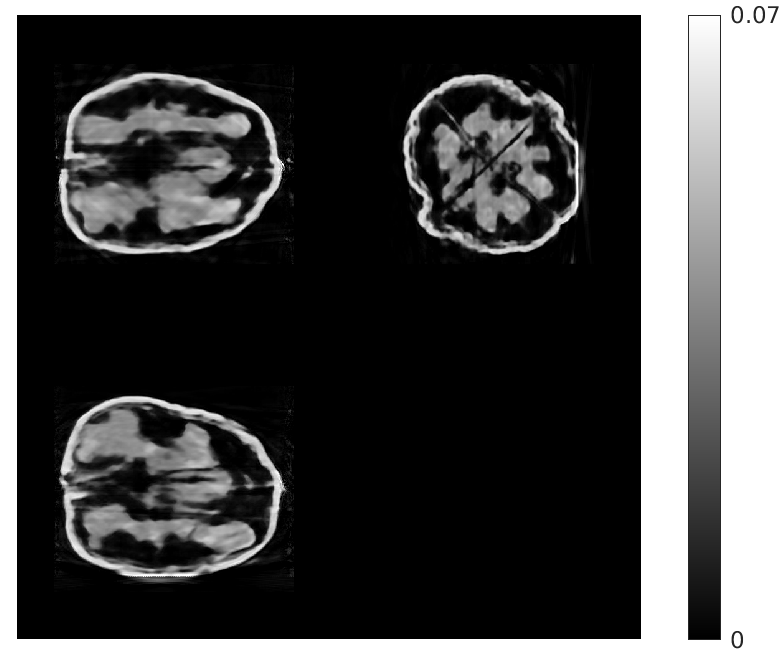}  \\
 (e) (0.40) \hspace{0.1in} & (f) (0.26) \hspace{0.1in} 
\end{tabular}
\caption{Comparison of the quality of reconstruction of our proposed algorithm (f) for walnut 1 (8 views) in \tref{table:comp_mae_hfen} to various reference methods. Each subfigure depicts slices through the center of the walnut volume in the sagittal, coronal and transverse orientations. The normalized mean absolute errors have also been shown underneath each subfigure. The central slices corresponding to the ground truth training walnut volume have also been shown in (b).}
\label{fig:recon:wal1}
\end{center}
\vspace{-0.1in}
\end{figure*}

\begin{figure*}
\begin{center}
\begin{tabular}{cc}
\textbf{Ground Truth Test Volume} & \\ 
\includegraphics[height=\hh]{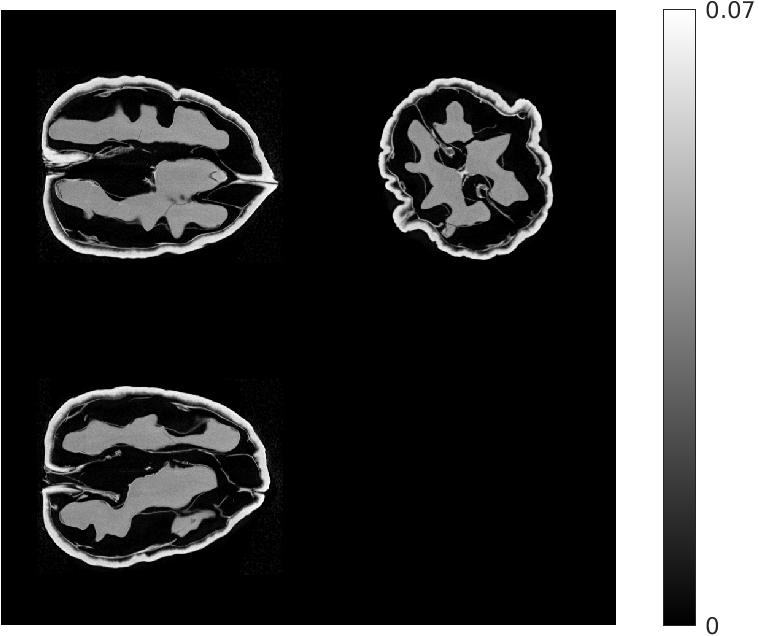} &  \\
\hspace{0.1in}(a) (NMAE)\hspace{0.1in} & \hspace{0.1in}\\
\textbf{FDK} & \textbf{EP Regularized Recon.} \\ 
 \includegraphics[height=\hh]{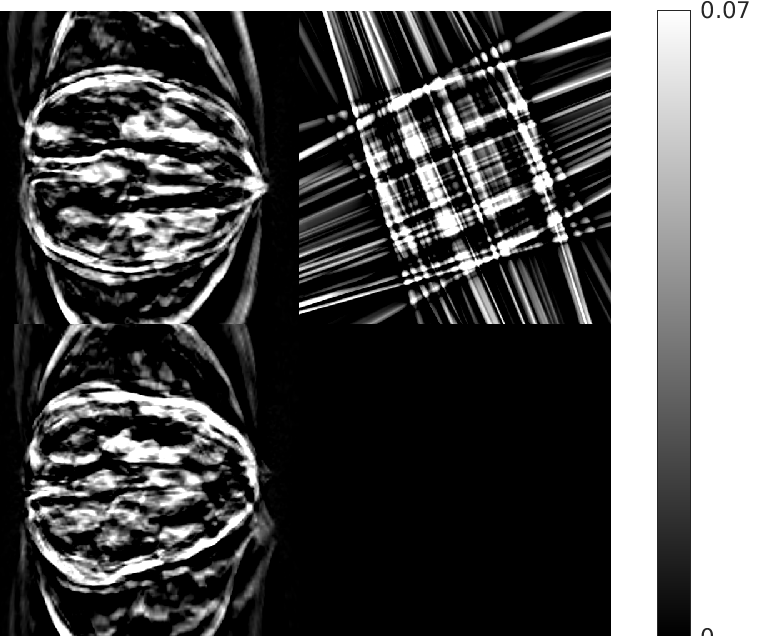}  & \includegraphics[height=\hh]{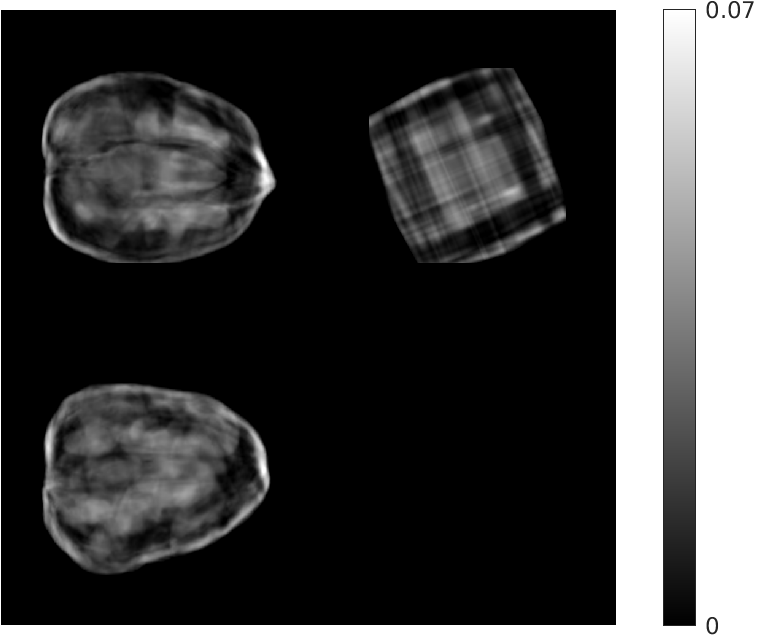}\\
 \hspace{0.1in}(b) (1.23) \hspace{0.1in} & \hspace{0.1in}(c) (0.65)\hspace{0.1in }\\
\textbf{Destreaking CNN} & \textbf{Proposed}\\
 \includegraphics[height=\hh]{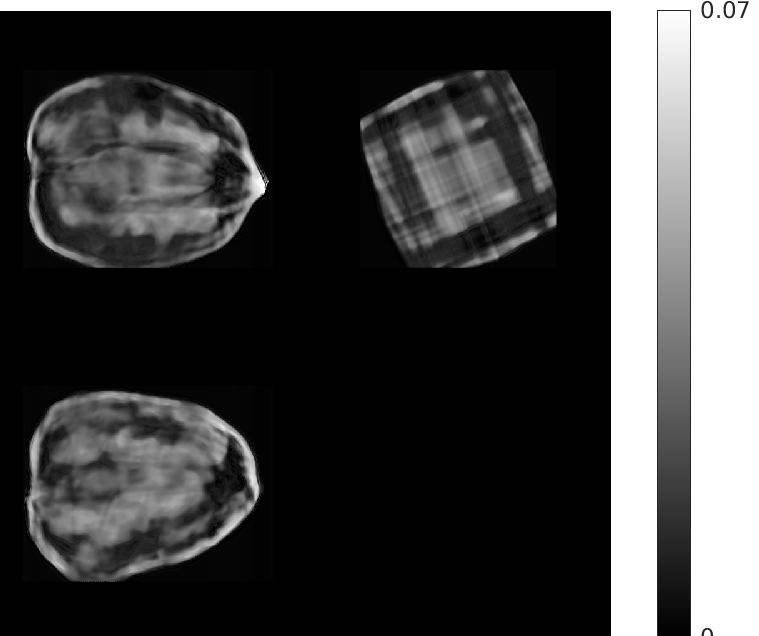}  & \includegraphics[height=\hh]{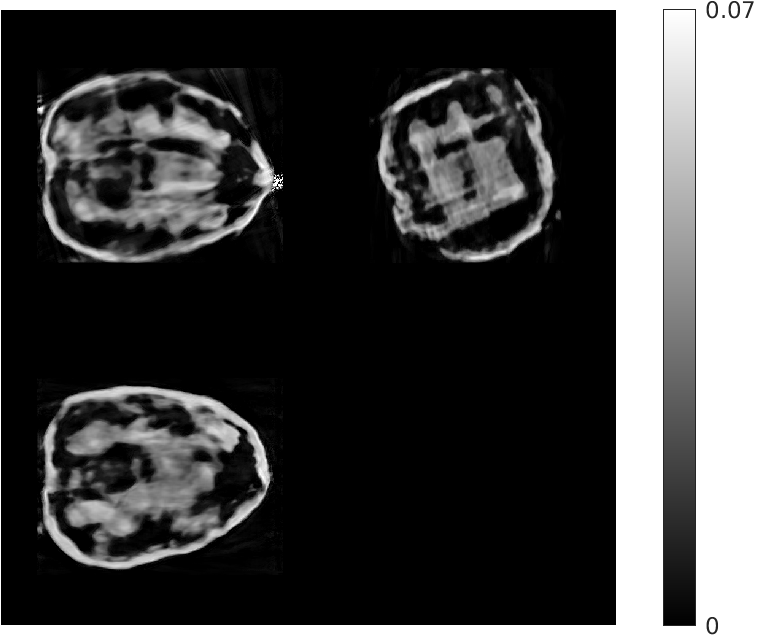}  \\
 (d) (0.63) \hspace{0.1in} & (e) (0.45) \hspace{0.1in} 
\end{tabular}
\caption{Comparison of the quality of reconstruction of our proposed algorithm (e) for walnut 2 (4 views) in \tref{table:comp_mae_hfen} to various reference methods. Each subfigure depicts slices through the center of the walnut volume in the sagittal, coronal and transverse orientations. The normalized mean absolute errors have also been shown underneath each subfigure.}
\label{fig:recon:wal2_4}
\end{center}
\vspace{-0.1in}
\end{figure*}

\edd{
\fref{fig:recon:wal2_4}
compares the reconstructions from 4 views.
The quality of the reconstruction is poorer compared to that using 8 views,
though
the proposed approach still visibly outperforms the other methods.
}

\fref{fig:prog_stage:wal2}
shows the results of the intermediate steps
of the first stage of our reconstruction for walnut 2 (8 views)
in the  test dataset.
While the patch-based destreaking compensates
for the blurring introduced by the EP-regularized reconstruction
and `fills-in' details in the reconstructed volume,
the data-consistency plays a key role in mitigating hallucinations introduced by the CNN,
and reinforces image features that are consistent with the acquired measurements.

\newcommand{\h}{1.8in}

\begin{figure*}[h!]
\begin{center}
\begin{tabular}{ccc}
\textbf{Input EP Recon.} & \textbf{Post CNN-based Destreaking} & \textbf{Post Data-consistency}  \\ 
 \includegraphics[height=\h]{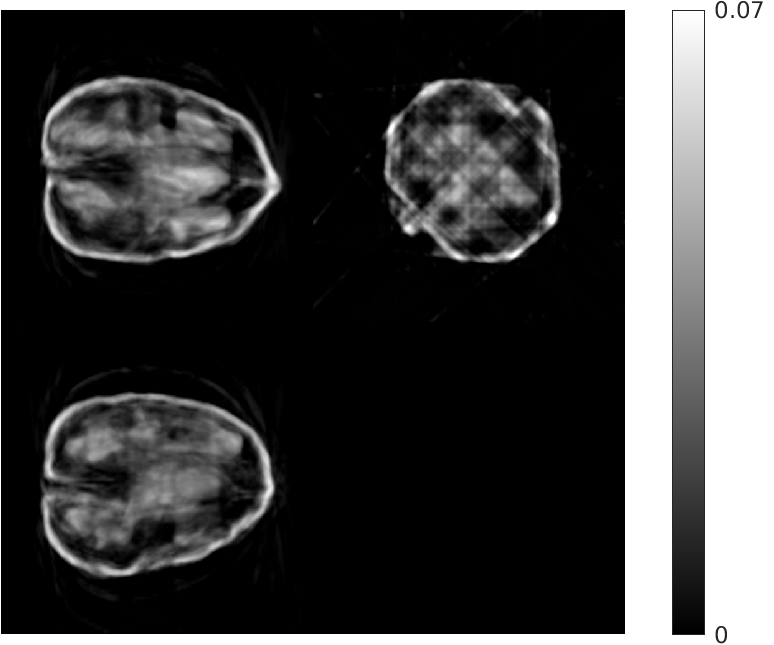}  & \includegraphics[height=\h]{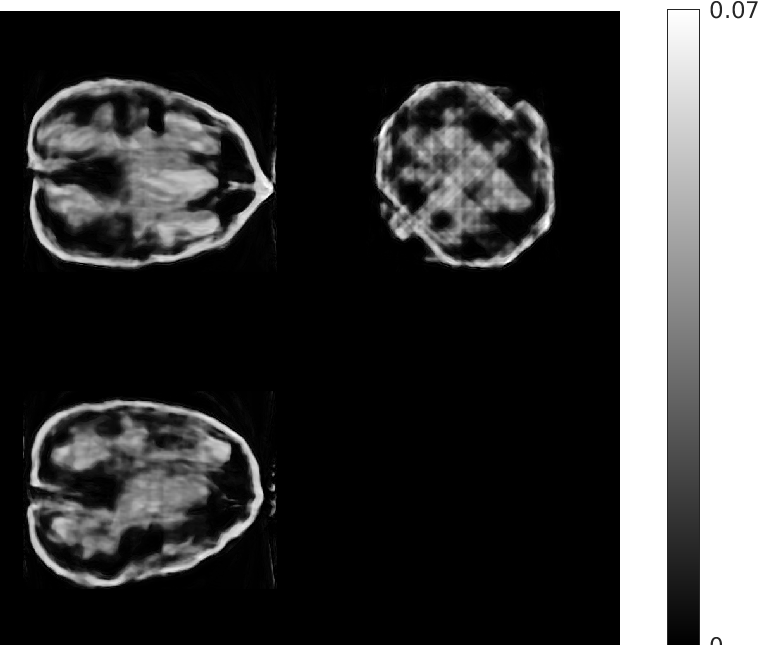} & \includegraphics[height=\h]{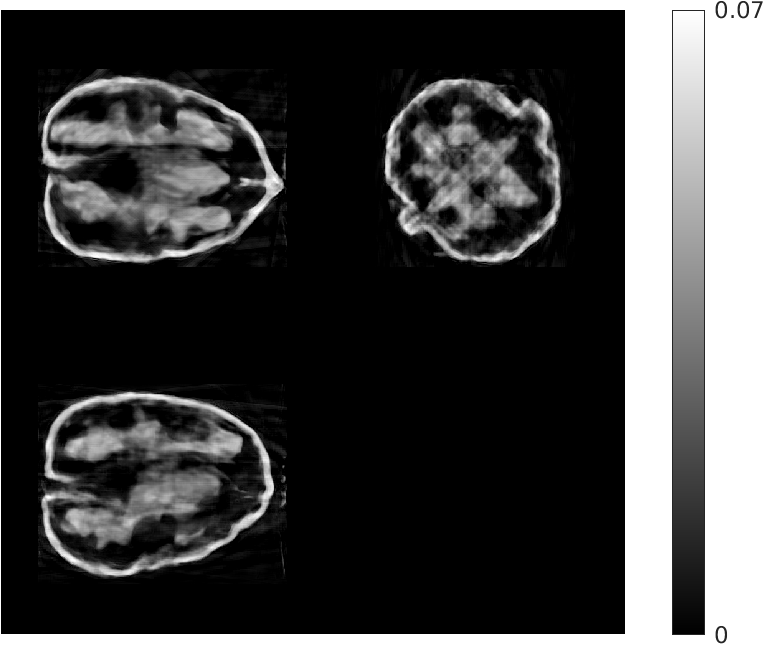} \\
 (NMAE:0.45) & (NMAE:0.38) & (NMAE:0.32)
\end{tabular}
\caption{Central slices corresponding to all three orientations of test walnut 2 (8 views) after various steps during a single stage (Stage 1) of our proposed algorithm.}
\label{fig:prog_stage:wal2}
\end{center}
\vspace{-0.1in}
\end{figure*}

\begin{figure*}
\begin{center}
\begin{tabular}{cc}
\textbf{Stage 1} & \textbf{Stage 2} \\ 
 \includegraphics[height=\hh]{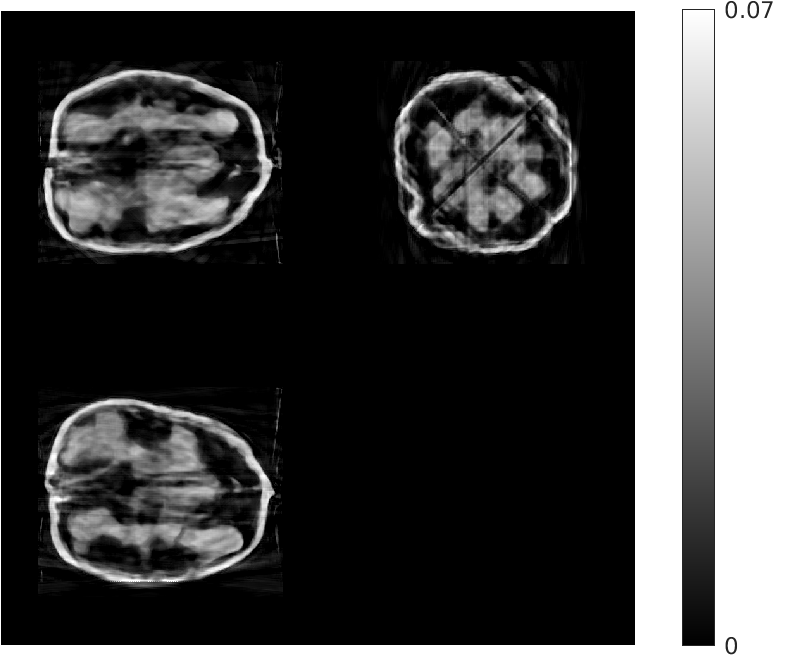}  & \includegraphics[height=\hh]{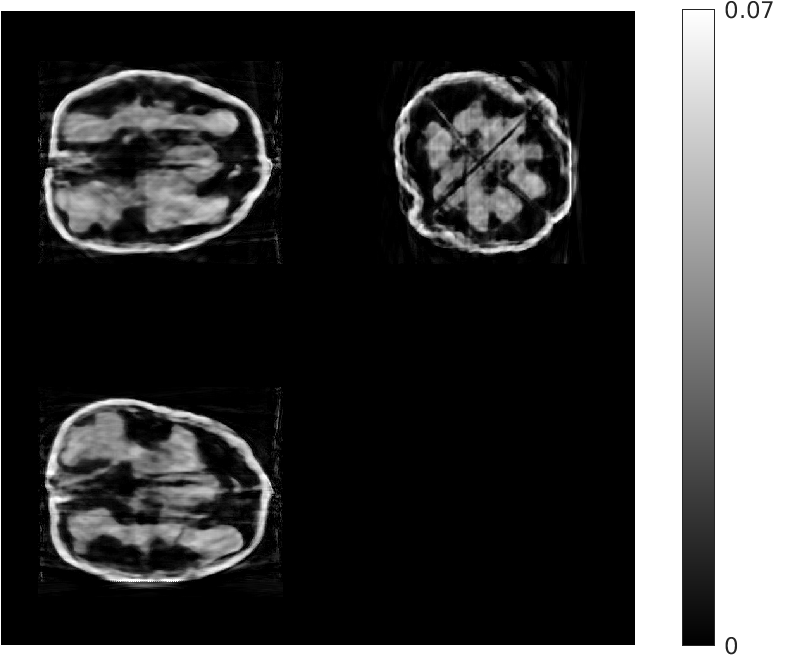}\\
 \hspace{0.1in}(a) (MAE: 0.32) \hspace{0.1in} & \hspace{0.1in}(b) (MAE: 0.29)\hspace{0.1in }\\
\textbf{Stage 3} & \textbf{Stage 4}\\
 \includegraphics[height=\hh]{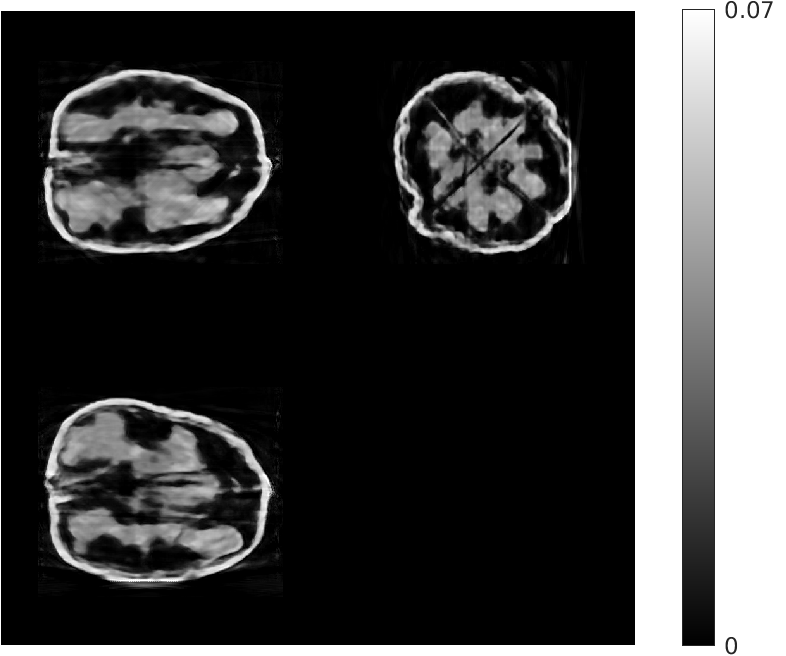}  & \includegraphics[height=\hh]{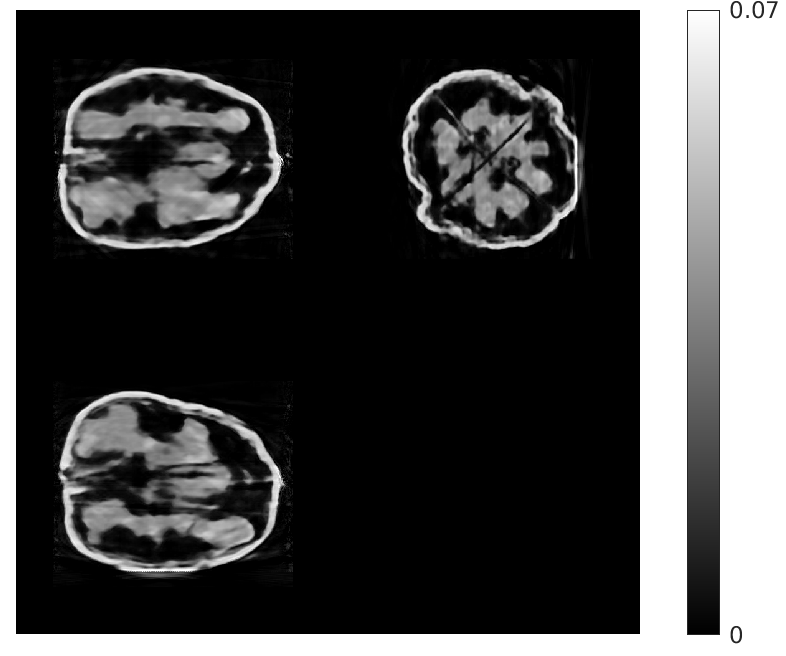}  \\
 (c) (MAE: 0.27) \hspace{0.1in} & (d) (MAE: 0.26) \hspace{0.1in} 
\end{tabular}
\caption{Central slices through the reconstructions of walnut 1 (8 views) in all three orientations at the end of each of the 4 stages of our algorithm (post data-consistency). It is evident that the quality of the reconstruction progressively improves across the stages. The normalized mean absolute error (NMAE) for each figure is also provided underneath  }
\label{fig:prog:wal1}
\end{center}
\vspace{-0.1in}
\end{figure*}

\fref{fig:prog:wal1} shows walnut 1 from our test dataset
being progressively reconstructed from 8 projections
across the stages of our algorithm;
as the stages progress,
more features are restored in the reconstructed walnut,
until the improvements become incremental.
The residual streaking artifacts outside the walnut
are mitigated in the reconstructions from the third and fourth stages.

\begin{figure*}[h!]
\begin{center}
\begin{tabular}{cc}
 \includegraphics[height=\h]{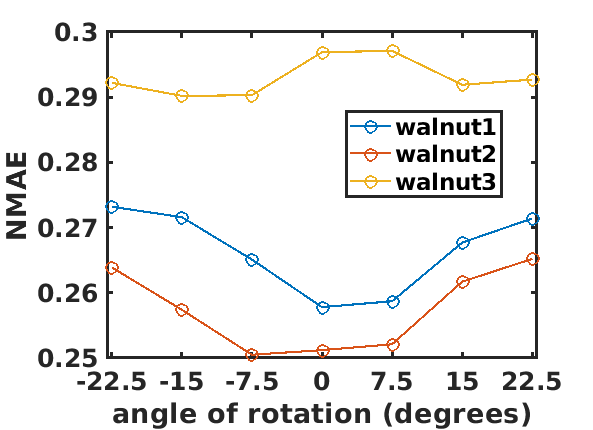} & \includegraphics[height=\h]{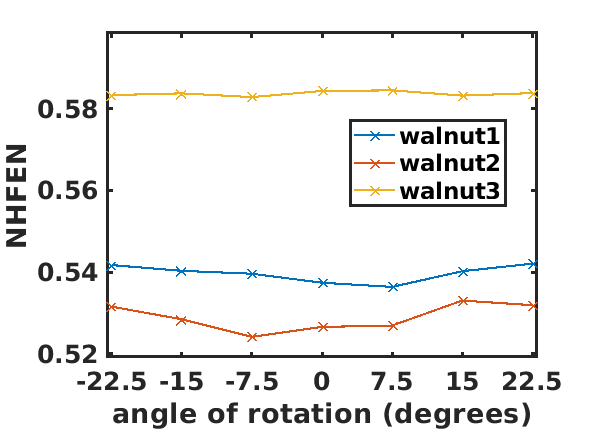} \\
(a) & (b)\\
\includegraphics[height=\h]{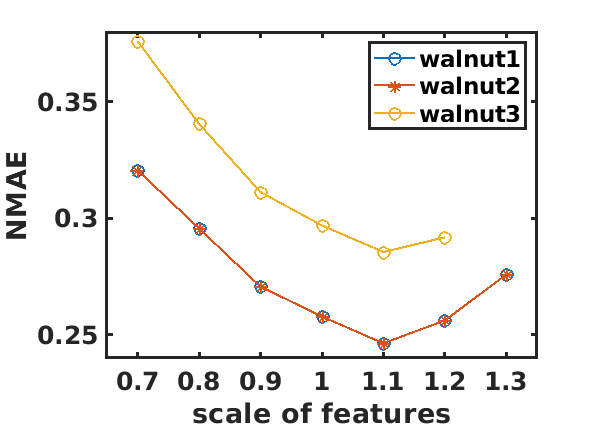} & \includegraphics[height=\h]{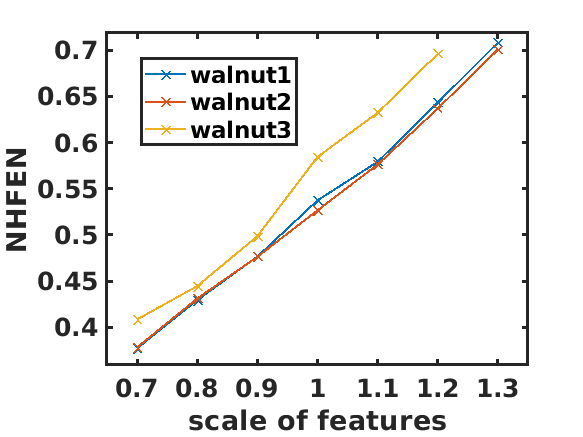}\\
(c) & (d)
\end{tabular}
\caption{(a) and (b) show the effects of rotating the test walnuts (by changing the angle of acquired projections) on the quality of the reconstructions for 8 acquired views, while (c) and (d) capture the effect of varying the scale of features of the test walnut compared to the scale used in training. (a) and (c) depict the normalized MAE metric while (b) and (d) depict the normalized HFEN error metric for respective experiments.}
\label{fig:rot:wal2}
\end{center}
\vspace{-0.1in}
\end{figure*}

\fref{fig:rot:wal2} (a) and (b) show how the quality of the reconstructions from 8 acquired views using our method varies when the 
orientation of the test walnuts is changed,
in comparison to training-time.
This is achieved by changing the position of the acquired (equidistant) projections,
as this is akin to rotating the test walnut.
For this purpose,
essentially the position of the 
first acquired projection is shifted by a specified angle. For the three test 
walnuts shown in the figure, the angle of rotation was changed between +22.5\degree 
and -22.5\degree in intervals of 7.5\degree. Both the normalized mean absolute error (NMAE) and the 
 normalized high-frequency error norm (NHFEN) were used as quality metrics in this experiment. 
Our method seems to be fairly robust to rotated test-data.

We also studied the effect of varying the scale of the features of  test walnuts on 
reconstructions from our method (from 8 projections), which was still trained on a single scale. 
For this purpose, we used the three walnuts which were used in the previous experiment.
From \fref{fig:rot:wal2} (c) and (d), it is evident that while there are differences
in the NMAE and NHFEN across scales, our method holds up reasonably well across 
multi-scale test data. 
Particularly, we notice that while higher-frequency 
features are better reconstructed when the test walnuts are scaled down
, the 
best NMAE is observed when the scale of the test walnut is 1.1 times its 
original scale.
We surmise that this happens because at a scale of 1.1, the scale of features of the
test walnuts matches that of the training walnut very well. 
(We excluded scale 1.3 for walnut 3
because at that scale,
the walnut exceeded the $501\times 501\times 501$ voxel grid.)


%% file: s,disc.tex
Our observations in Section~\ref{results} indicate that the proposed physics-aware learning-based approach
for limited-view CBCT reconstruction is able to improve upon the quality of reconstructions yielded not 
only by the FDK algorithm, but also traditional prior-based iterative reconstruction, which serves a 
crucial role in initializing our algorithm, and ensures that the subsequent CNN-based destreaking is 
afforded a reasonable mapping to learn when trained with ground truth images as targets.
Furthermore, in \fref{fig:prog_stage:wal2}, we also demonstrated the importance of data-consistency when 
reconstructing from limited measurements. The data consistency update could easily be the most crucial step in 
our algorithm because it corrects for hallucinations introduced by the CNN-based destreaking step, which are very likely to occur given the limited availability of training data.
This allows our algorithm to be repeated for several stages, which is key to the improvement provided by the proposed method over a simple image domain CNN-based denoiser. 
The generalizability of our approach is evident from the consistent improvements yielded by our approach in \tref{table:comp_mae_hfen}, which is a consequence of 
using subvolumes or patches in training our method as well as the relative shallowness of the CNNs used therein. These measures effectively reduce the chances of overfitting when training on extremely limited full-view data. 
The chosen patch size effectively allowed the learned destreaking CNN to utilize three-dimensional context while also reducing: (1) the 
total number of overlapping subvolumes that are forward propagated through the destreaking CNN at test time and (2) the memory requirements for training the network.
The hybrid 3D to 2D mapping also plays an important role by reducing the computational and time demands of our algorithm, and by removing any chances of artifacts that may be introduced during (conventional overlapping) patch aggregation. 

We noted that the final reconstruction quality for some test walnuts was better than others, and also that the final reconstruction quality from our algorithm depended on the quality 
of the initial EP reconstruction. We think this may be because the acquired views (equidistant over $360^\circ$)
may not be the best choice
for higher fidelity reconstructions for some walnuts,
and because these walnuts may be less similar to the training walnut. 

Regarding the performance of our reconstructions in terms of the NHFEN metric, 
we observe a lack of sensitivity with respect to other compared techniques for both
8 and 4 views. We surmise that this is because neither of the networks was trained for 
improved NHFEN, i.e., a NHFEN term was not part of the training loss.

While the experiments with rotated projections and varying scale of features 
showcased some robustness of our proposed approach to both rotation and 
up/downscaling of features-- it also brought to light the necessity for data 
augmentation for improved performance. 
The consistent degradation of the high frequency features (observed through the NHFEN metric) in reconstructions with 
increase in scale was interesting to note, and its contrast with the trends in NMAE
points towards a sharpness/fidelity trade-off that may need more investigation. 

However, it is key to realize that the improved performance in the NHFEN metric
at smaller scales using our method is likely due to fewer edges that need to 
be reproduced in the image, and does not suggest that reconstructions of 
increasingly sharper quality may be obtained by continually shrinking the scale of walnuts. 

%% file: s,conc.tex
This paper developed a method to provide high quality reconstructions
from extremely limited CBCT projections and scarce training data. 
The key features of our approach
were the multi-stage approach of alternating
between learning-based destreaking and data consistency
and the use of subvolume-based learning
and shallower (adversarially trained) CNNs to combat over-fitting. 
In the future, we will focus on extending our method to dynamic imaging applications in CT,
as well as being able to jointly segment and reconstruct three-dimensional objects.
Towards this end, we are also interested in finding a better metric
than the mean absolute error
to assess the fidelity and quality of our reconstructions for specific tasks.

\textsl{
The code for reproducing the results in this paper
will be available on github
after the paper is accepted.
}